\documentclass{article}
\usepackage{spconf,amsmath,graphicx}
\usepackage{amssymb}
\usepackage{booktabs}
\usepackage{multicol}
\usepackage{multirow}
\usepackage{makecell}
\usepackage{setspace}
\usepackage{bm}
\usepackage{caption}
\usepackage{pifont}
\usepackage[hidelinks]{hyperref}
\title{TALDS-Net: Task-Aware Adaptive Local Descriptors Selection for Few-shot Image Classification}
\name{{Qian Qiao, Yu Xie\dag, Ziyin Zeng, Fanzhang Li\sthanks{Corresponding author. \dag Yu Xie does equal Contribution.
} }}
\address{School of Computer Science and Technology, Soochow University, China}
\begin{document}
\ninept
\maketitle

\begin{abstract}
Few-shot image classification aims to classify images from unseen novel classes with few samples. Recent works demonstrate that deep local descriptors exhibit enhanced representational capabilities compared to image-level features. However, most existing methods solely rely on either employing all local descriptors or directly utilizing partial descriptors, potentially resulting in the loss of crucial information. Moreover, these methods primarily emphasize the selection of query descriptors while overlooking support descriptors. In this paper, we propose a novel Task-Aware Adaptive Local Descriptors Selection Network (TALDS-Net), which exhibits the capacity for adaptive selection of task-aware support descriptors and query descriptors. Specifically, we compare the similarity of each local support descriptor with other local support descriptors to obtain the optimal support descriptor subset and then compare the query descriptors with the optimal support subset to obtain discriminative query descriptors. Extensive experiments demonstrate that our TALDS-Net outperforms state-of-the-art methods on both general and fine-grained datasets.
\end{abstract}
\section{Introduction}
\label{sec:intro}
In data-scarce tasks (\emph{e.g.}, medical images), generic deep learning methods fail to achieve outstanding performance. To cope with these challenges, researchers have proposed few-shot learning to enable models to rapidly adapt to new tasks using only a limited amount of training data. These methods can be classified into three categories: optimization-based methods\cite{finn2017model,antoniou2018train,ravi2016optimization}, mertic-based methods\cite{snell2017prototypical,sung2018learning,vinyals2016matching} and data augmentation based methods\cite{antoniou2017data,zhang2018metagan,schwartz2018delta,xian2019f,chen2019multi,yang2021free}. Optimization-based methods aim to learn a globally optimal set of initial parameters, allowing the model to quickly adapt to new tasks through stochastic gradient descent based on these parameters. Data augmentation based methods leverage generative models or enhanced feature spaces to ultimately facilitate few-shot learning. Metric-based methods focus on learning an appropriate deep embedding space for measuring query samples against all support classes.

This work is based on metric-based few-shot learning methods. Metric learning approaches primarily focus on concept representation or relationship measurement. Early metric-based few-shot classification methods commonly leverage image-level representations\cite{koch2015siamese,vinyals2016matching,snell2017prototypical}. While these methods excel with large-scale data, they struggle in few-shot tasks due to the difficulty of obtaining discriminative features.
\par
Recent studies \cite{li2019revisiting,liu2022dmn4,zhang2020deepemd,dong2021learning} have demonstrated that deep local descriptors provide superior representation compared with image-level features. The early work \cite{sung2018learning} proposes a relation network, implicitly utilizing local descriptors. DN4 \cite{li2019revisiting} directly selects $k$ support local descriptors for each query local descriptor using the $k$-nearest neighbor algorithm, approximates the comparison between query samples and support classes based on cosine similarity distance. 
Based on DN4, both \cite{dong2021learning} and \cite{liu2022dmn4} proposed ATL-Net and DMN4, respectively. ATL-Net adaptively selects important local descriptors for classification by measuring the relationships between each query local descriptor and all support classes. DMN4 explicitly selects query descriptors most relevant to each task by establishing Mutual Nearest Neighbor (MNN) relationships.
However, DN4\cite{li2019revisiting} directly utilizes all query local descriptors, including potentially task-irrelevant ones such as background and redundant descriptors. DMN4\cite{liu2022dmn4} assumes that not all query descriptors are valuable for classification and applies direct filtering rules to select local descriptors, which could result in the loss of essential local descriptors.
\par
Motivated by the above observation, we propose a Task-Aware Adaptive local descriptors selection Network (TALDS-Net) to adaptively select query descriptors and support descriptors. Based on human learning and recognition processes, when humans encounter new knowledge and use it to identify unfamiliar images, they only need to know a few critical characteristics of a particular class of objects they are already familiar with. Humans compare these critical characteristics with those of unfamiliar classes to classify them. TALDS-Net achieves this by employing two simple learnable modules $\mathcal{F}_\Gamma$ and $\mathcal{F}_\Psi$(as shown in 2.3) to predict thresholds adaptively and then utilizes these learned thresholds and attention maps to select the most discriminative descriptors.  Initially, the attention maps filter the local descriptors of various classes within the support set, selecting a discriminative subset of support descriptors. Subsequently, the selected subset of support descriptors is utilized to identify the most discriminative query descriptors. Experimental results on both general and fine-grained datasets demonstrate that our method outperforms other state-of-the-art methods.

\begin{figure*}[!htb]
\setlength{\abovecaptionskip}{0.5cm}
\setlength{\belowcaptionskip}{-0.5cm}
\centerline{\includegraphics[width=1.0\linewidth]{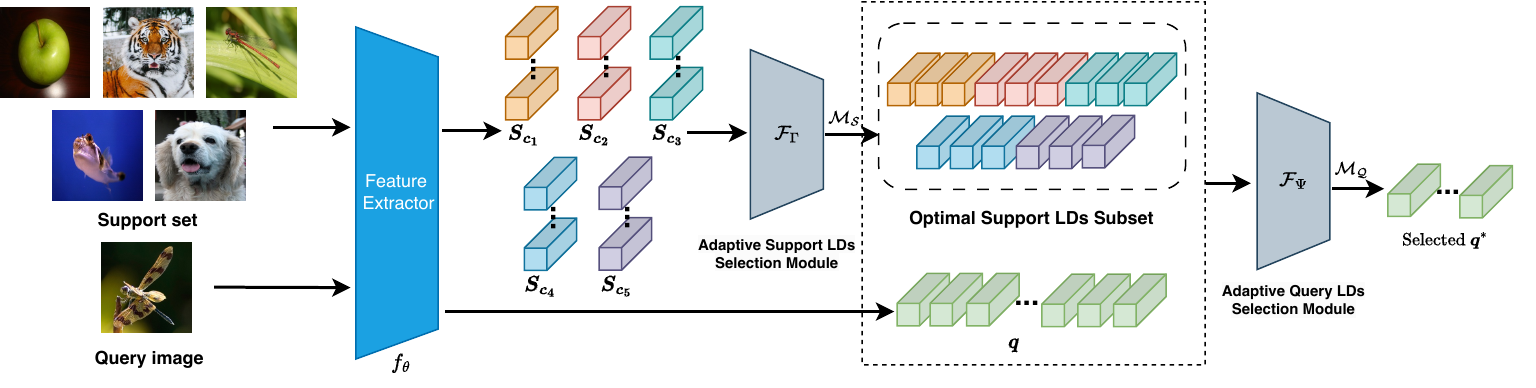}}
\caption{The overall architecture of the proposed TALDS-Net. TALDS-Net extracts descriptors from support and query images using $f_\theta$. Support descriptors are selected into an optimal subset through the adaptive selection module $\mathcal{F}_\Gamma$. Subsequently, query descriptors are adaptively selected through $\mathcal{F}_{\Psi}$.}
\label{fig:1}       
\end{figure*}
\par
In summary, our main contributions are three folds:

\textbf{(1)}We propose a novel Task-Aware Adaptive Local Descriptors Selection Network (TALDS-Net), which adaptively selects task-aware support and query descriptors. 

\textbf{(2)}We introduce the concept of discriminative support descriptors for the first time and analyze the advantages of using this method compared to directly using all query local descriptors or selecting a portion of local descriptors.

\textbf{(3)}Extensive experimental results demonstrate that TALDS-Net outperforms state-of-the-art methods on multiple general and fine-grained datasets.
\section{METHOD}
\label{sec:method}
The overall architecture is illustrated in Fig \ref{fig:1}, TALDS-Net consists of a feature extractor $f_{\theta}$, a support descriptor selection module $\mathcal{F}_\Gamma$, and a query descriptor selection module $\mathcal{F}_\Psi$. Specifically, TALDS-Net extracts descriptors from support and query images using $f_\theta$. $\mathcal{F}_\Gamma$ computes the support descriptor attention map $\mathcal{M^{\prime}}_A$ for adaptive support descriptor selection, while $\mathcal{F}_\Psi$ computes the query descriptor attention map $\mathcal{M}_A$ for adaptive query descriptor selection. Different backbones can be used for $f_\theta$.
\subsection{Problem Definition}
In this paper, we follow the common settings of few-shot learning methods. Typically, there are three datasets: a training dataset $A$, a support dataset $S$, and a query dataset $Q$. $S$ comprises $N$ classes, each containing $K$ labeled samples. $S$ and $A$ have different label spaces with no intersection. Our goal is to classify query samples $q\in Q$ into one of the $N$ support classes, which is referred to as the $N$-way $K$-shot task. To achieve this goal, we employ the episodic training mechanism\cite{vinyals2016matching} using an auxiliary set $A$ to learn transferable knowledge. We construct numerous $N$-way $K$-shot tasks from $A$ during the training phase, where each task consists of an auxiliary support set $A_S$ and an auxiliary query set $A_Q$. In the training stage, tens of thousands of tasks are fed into the model, encouraging it to learn transferable knowledge that can be applied to new $N$-way $K$-shot tasks involving unseen classes. In the testing stage, the trained model, based on $S$, is employed to classify the images in $Q$.
\subsection{Image Representation Based on Local Descriptors}
We obtain a three-dimensional feature representation $f_{\theta}(X)\in \mathbb{R}^{h\times w\times d}$ for a given image $X$ using the embedding module $f_{\theta}(\cdot)$. Similar to previous Methods \cite{li2019revisiting,liu2022dmn4}, we consider it as a set of $m$ ($m = h \times w$) $d$-dimensional descriptors as: 
\begin{equation}
    \begin{split}
        f_{\theta}(X)=[x_1,x_2,\cdots,x_m]\in \mathbb{R}^{m \times d}
    \end{split}
\end{equation}
where $x_i$ represents the $i$-th deep local descriptor. In each episode, when using shallower embedding modules (\emph{e.g.}, Conv-4), each support class is represented in its original form. When using deeper embedding modules (\emph{e.g.}, ResNet-12), each support class is represented by the empirical mean of its support descriptors. 
\begin{table*}

\caption{5-way 1-shot and 5-shot classification accuracies on miniImageNet and tieredImageNet datasets using Conv-4 and ResNet-12 backbones with $95\%$ confidence intervals. All the results of comparative methods are from the exiting literature ('-' not reported).}
\label{tab1}
\resizebox{\linewidth}{!}{
\begin{tabular}{lcccccccc}
\toprule
\multicolumn{1}{c}{\multirow{3}{*}{\textbf{Method}}} & \multicolumn{4}{c}{\textbf{Conv-4}}                        & \multicolumn{4}{c}{\textbf{ResNet-12}} \\
\multicolumn{1}{c}{} &
  \multicolumn{2}{c}{miniImageNet} &
  \multicolumn{2}{c}{tieredImageNet} &
  \multicolumn{2}{c}{miniImageNet} &
  \multicolumn{2}{c}{tieredImageNet} \\ 
  \cmidrule(r){2-3} \cmidrule(r){4-5} \cmidrule(r){6-7} \cmidrule(r){8-9}
\multicolumn{1}{c}{}& 1-shot & 5-shot & 1-shot & 5-shot & 1-shot & 5-shot & 1-shot & 5-shot \\ \hline
MatchingNet\cite{vinyals2016matching}   & 43.56 $\pm$ 0.84 & 55.31$\pm$0.73 & - & - & 63.08$\pm$0.20 & 75.99$\pm$0.15 & 68.50$\pm$0.92 & 80.60$\pm$0.71 \\
ProtoNet\cite{snell2017prototypical}     & 51.20$\pm$0.26 & 68.94$\pm$0.78 & 53.45$\pm$0.15 & 72.32$\pm$0.57 & 62.33$\pm$0.12 & 80.88$\pm$0.41 & 68.40$\pm$0.14 & 84.06$\pm$0.26 \\
RelationNet\cite{sung2018learning}  & 50.44$\pm$0.82 & 65.32$\pm$0.70 & 54.48$\pm$0.93 & 71.31$\pm$0.78 & 60.97  & 75.12   & 64.71 & 78.41 \\
MetaOptNet\cite{lee2019meta}   & 52.87$\pm$0.57 & 68.76$\pm$0.48 & 54.71$\pm$0.67 & 71.79$\pm$0.59 & 62.64$\pm$0.61 & 78.63$\pm$0.46 & 65.99$\pm$0.72 & 81.56$\pm$0.53 \\
CovaMNet\cite{li2019distribution}     & 51.19$\pm$0.76 & 67.65$\pm$0.63 & 54.98$\pm$0.90 & 71.51$\pm$0.75 & -  & -  & -   & -  \\
DN4\cite{li2019distribution}   & 51.24$\pm$0.74 & 71.02$\pm$0.64 & 52.89$\pm$0.23 & 73.36$\pm$0.73 & 65.35 & 81.10 & 69.60 & 83.41 \\
DeepEMD\cite{zhang2020deepemd}   & 51.72$\pm$0.20 & 65.10$\pm$0.39 & 51.22$\pm$0.14 & 65.81$\pm$0.68 & 65.91$\pm$0.82 & 82.41$\pm$0.56 & 71.16$\pm$0.87 & 86.03$\pm$0.58 \\
RFS-Simple\cite{tian2020rethinking} & 55.25$\pm$0.58 & 71.56$\pm$0.52 & 56.18$\pm$0.70 & 72.99$\pm$0.55 & 62.02$\pm$0.63 & 79.64$\pm$0.44 & 69.74$\pm$0.72 & 84.41$\pm$0.55 \\
RFS-Distill\cite{tian2020rethinking}  & 55.88$\pm$0.59 & 71.65$\pm$0.51 & 56.76$\pm$0.68 & 73.21$\pm$0.54 & 64.82$\pm$0.60 & 82.14$\pm$0.43 & 71.52$\pm$0.69 & 86.03$\pm$0.49 \\
ATL-Net\cite{dong2021learning}   & 54.30$\pm$0.76 & 73.22$\pm$0.63 & - & -  & -  & -  & - & -  \\
FRN\cite{wertheimer2021few}  & 54.87  & 71.56 & 55.54 & 74.68 & 66.45$\pm$0.19 & 82.83$\pm$0.13 & \textbf{72.06$\pm$0.22} & \textbf{86.89$\pm$0.14} \\
DMN4\cite{liu2022dmn4} & 55.77  & 74.22 & 56.99 & 74.13 & 66.58  & 83.52  & 72.10 & 85.72  \\ \hline
\textbf{TALDS-Net(ours)} &
  \multicolumn{1}{l}{\textbf{56.78$\pm$0.37}} &
  \multicolumn{1}{l}{\textbf{74.63$\pm$0.13}} &
  \multicolumn{1}{l}{\textbf{57.54$\pm$0.71}} &
  \multicolumn{1}{l}{\textbf{75.79$\pm$0.43}} &
  \multicolumn{1}{l}{\textbf{67.89$\pm$0.20}} &
  \multicolumn{1}{l}{\textbf{84.31$\pm$0.44}} &
  \multicolumn{1}{l}{71.34$\pm$0.32} &
  \multicolumn{1}{l}{86.12$\pm$0.33} \\
  \bottomrule
\end{tabular}}
\end{table*}
\subsection{Local Descriptors Selection}
\textbf{Support Local Descriptors Selection.} As mentioned above, we need to select local descriptors relevant to the task. Given an image $X^j$ in support class $j$, it is embedded as $f_{\theta}(X^j)=[x_1^j,\cdots,x_m^j]\in \mathbb{R}^{m\times d}$. For each support descriptor $x_i^{j}$, We find the $k$ nearest neighbor support descriptors in class $c$, denoted as $kNN(x_i^{j})=\{\hat 
x_1^c,\cdots,\hat x_k^c\}$, note that when the support descriptor $x_i^{j}$ is from class $c$ (\emph{i.e.}, $j=c$), $x_i^j$ itself is excluded from the search for the $k$ nearest neighbor descriptors. Then, we calculate the sum of cosine distances between $x_i^j$ and each $\hat x^c$ to represent the similarity between the support descriptor $x_i^j$ and the support class $c$: 
 \begin{equation}
    \begin{split}
        \mathcal{\gamma}^{x_i^j}_{c}=\sum_{\hat{x}^c\in \text{kNN}(x_i^{j})}g(x_i^j,\hat{x}^c)
    \end{split}
\end{equation}
where $c\in \{1, 2, \cdots, N\}$ denotes support class, $g(\cdot)$ is a similarity metric, which is implemented as cosine similarity in this paper. Next, we use $softmax$ to normalize $\mathcal{\gamma}^{x_i^j}_{c}$ and calculate the discriminant score :
\begin{equation}
    \begin{split}
        \mathcal{R}^{x_i^j} = \max_c(\text{softmax}(\mathcal{\gamma}^{x_i^j}_{c})
    \end{split}
\end{equation}
Accordingly, we obtain discriminative scores $\mathcal{R}^{x_i^j}$ for each support descriptor. 
Inspired by \cite{dong2021learning}, we employ a learnable module $\mathcal{F}_\Gamma$ to learn a support descriptor attention map $\mathcal{M_S}$. And it can adaptively select support descriptors with discriminative qualities. Specifically, we use a multi-layer perceptron (MLP) denoted as $\mathcal{F}_{\Gamma}$ to adaptively predict the threshold $\mathcal{V}^*_{c^{\prime}}$. Taking the current support descriptor $x_i^j$ and all the remaining support descriptors $\hat L=\{x_i^{1}, \cdots, x_i^{j-1}, x_i^{j+1}, \cdots, x_i^{N}\}, i=\{1,\cdots,m\}$ as input. The threshold $\mathcal{V}^*_{c^{\prime}}$ is calculated as :
\begin{equation}
    \begin{split}
        \mathcal{V}^*_{c^{\prime}} = \sigma(\mathcal{F}_\Gamma(x_i^j, \hat L))
    \end{split}
\end{equation}
where $\sigma$ is a $sigmoid$ function. The calculation of values $\mathcal{M_S}$ in the final support descriptor attention map $\mathcal{M}_s$ is as follows:
\begin{equation}
    \begin{split}
        \mathcal{M}_s = {1}/{(1+\exp^{-\lambda_1(\mathcal{R}^{x_i^j}-\mathcal{V}^*_{c^{\prime}})})}
    \end{split}
\end{equation}
$\mathcal{M}_s$ is a variant of sigmoid function. Theoretically, when $\lambda_1$ is sufficiently large and $\mathcal{R}^{x_i^j}>\mathcal{V}^*_c$, $\mathcal{M}_s$ approximates $1$. Conversely, if $\mathcal{R}^{x_i^j}<\mathcal{V}^*_c$, $\mathcal{M}_s$ approximates $0$. 

And the similarity score between each image $X$ and each support class $c$ is $\text{score}(X,c)=\sum_{x_i^j\in f_{\theta}(X)} \mathcal{\gamma}^{x_i^j}_{c} \mathcal{M}_s$. Therefore, the support descriptor attention map can effectively select the support descriptors. 
We use Adam optimizer with a cross-entropy loss to train the network. 
In the end, we obtain the optimal subset of support descriptors $\boldsymbol{S^{*}}$ for each episode.

\textbf{Query Local Descriptors Selection.} 
In this section, our method is similar to the method used for support descriptors. Given a query image $X_q$ embedded as $f_{\theta}(X_q)=[x^q_1,\cdots,x^q_m]\in \mathbb{R}^{m\times d}$. First, we find the $k$ nearest neighbors of the query desciptors $x_i^q$ in the optimal support descriptor subset $\boldsymbol{S_c^*}$ for class $c$, denoted as $kNN(x^q_i)=\{s_1^c, s_2^c, \cdots, s_k^c\}$. Then, we calculate the similarity between each query descriptor $x_i^q$ and its $k$ nearest neighbors $s^c$. Similarly, the discriminative score for each query descriptor $x^q_i$ is calculated as :
\begin{equation}
    \begin{split}
        &\mathcal{R}^q = \max_c(\text{softmax}(\mathcal{\gamma}^{x^q_i})) \\&
        \mathcal{\gamma}^{x_i^q}=\sum_{s^c\in kNN(x_i^q)} g(x_i^q,s^c)
    \end{split}
\end{equation}
where $c \in \{1, 2, \cdots, N\}$ denotes support classes, $i \in \{1,\cdots,m\}$ denotes number of query descriptors, $g(\cdot)$ is a similarity metric, which is implemented as cosine similarity in this paper.. 

In prior work \cite{li2019distribution,liu2022dmn4}, there are typically two methods used to select query descriptors. One method is to directly filter descriptors using a fixed threshold $\mathcal{V}$, while the other is to select the top $\tau$ query descriptors. However, both of these methods have poor generalization. Then we employ the learnable module $\mathcal{F}_\Psi$ to learn the query descriptor attention map $\mathcal{M_Q}$, which can adaptively choose discriminative query descriptors. This method is different from \cite{dong2021learning} and support descriptors selection. Our threshold $\mathcal{V}^*_c$ is determined jointly by both the support descriptors and the query descriptors. So, taking the optimal support descriptor subset and query descriptors as input, the output $\mathcal{V}^*c$ is computed, and the final calculation for the values $\mathcal{M}_q$ in the query descriptor attention map $\mathcal{M_Q}$ is as follows:
\begin{equation}
    \begin{split}
        &\mathcal{M}_q = {1}/{(1+\exp^{-\lambda_2(\mathcal{R}_q-\mathcal{V}^*_{c})})}\\&
        \mathcal{V}^{*}_c = \sigma(\mathcal{F}_\Psi(\boldsymbol{S^*},\boldsymbol{q}))
    \end{split}
\end{equation}
where $\boldsymbol{S^*}$ denotes the optimal support descriptor subset, $\boldsymbol{q}$ denotes the query descriptor.
Therefore, we can select query descriptors using the attention map $\mathcal{M_Q}$. According, the similarity score between each query image $X_q$ and each support class $c$ is calculated as follow:
\begin{equation}
    \begin{split}
        \text{score}(X_q,c)=\sum_{x^q_i\in f_{\theta}(X_q)} \mathcal{\gamma}^{x^q_i} \mathcal{M}_q
    \end{split}
\end{equation}
And the loss function can be defined as follows:
\begin{equation}
    \begin{split}
        &\mathcal{J}({\phi})=-\frac{1}{|A_Q|}\sum_{X_q\in A_Q}\sum_{c=1}^N y\log p_\phi\left(y=c \mid X_q\right)\\ &
        p_\phi\left(y=c \mid X_q\right)=\frac{\exp \left(\text{score}(X_q,c)\right)}{\sum_{c^{\prime}=1}^N \exp \left(\text{score}(X_q,c^{\prime})\right)}
    \end{split}
\end{equation}
where $y$ is the label of the query image $X_q$ and $c^{\prime}\neq c$.
\section{EXPERIMENTS}

\subsection{Datasets and Implementation Details}
\textbf{miniImageNet}\cite{vinyals2016matching}, a subset of ImageNet \cite{deng2009imagenet}, includes 100 classes, with each class having 600 images sized at 84x84 pixels. The dataset is divided into $64$ classes for training, $16$ classes for validation, and $20$ classes for testing.
\textbf{tieredImageNet}\cite{ren2018meta}, another subset of ImageNet, includes $608$ classes, with each class comprising $1281$ images. These $608$ classes are split into $351$ for training, $97$ for validation, and $160$ for testing.
\textbf{CUB-200}\cite{wah2011caltech} is a fine-grained dataset with $11788$ bird images, representinf $200$ different bird species. We select $100$ classes for training, $50$ classes for validation, and $50$ classes for testing. All images are resized to match the dimensions of miniImageNet, which are $84\times 84$ pixels.


\textbf{Network architecture.} Following previous works\cite{li2019distribution,li2019revisiting,dong2021learning}, we use Conv-4 and ResNet-12 as feature extraction networks $f_{\theta}$. The $f_{\theta}$ using Conv-4 comprises four convolutional blocks, each containing a convolutional layer, batch normalization layer, and Leaky ReLU layer. For the first two convolutional blocks, an additional $2\times 2$ max-pooling layer is attached to each of them. 
Conv-4 generates feature maps of size $19\times 19 \times 64$ for $84\times 84$ images, which means there are $361$ deep local descriptors of $64$ dimensions. 
The $f_{\theta}$ using ResNet-12 comprises four residual blocks, where each block contains three convolutional layers with $3\times 3$ kernels, three batch normalization layers, three Leaky ReLU layers, and a $2\times 2$ max-pooling layer. 
ResNet-12 generates feature maps of size $5\times 5 \times 640$ for $84\times 84$ images, resulting in $25$ deep local descriptors of $640$ dimensions. 
Both the MLP modules $\mathcal{F}_\Gamma$ and $\mathcal{F}_\Psi$ consists of two fully connected layers, a Leaky ReLU layer and a sigmoid function. The local descriptors are mapped to a subspace using a transformation layer $f\phi$, comprising a $1\times 1$ convolutional layer, a batch normalization layer, and a Leaky ReLU layer.

\textbf{Training and evaluation details.} During the meta-training phase, following the settings in \cite{liu2022dmn4}, for Conv-4, we trained from scratch for 30 epochs using the Adam optimizer with learning rate of $1e-3$, decaying by a factor of $0.1$ every $10$ epochs. the value of nearest neighbor $k$ is 1. For ResNet-12, we pre-trained it and then fine-tuned it for 40 epochs using momentum SGD with an initial learning rate of $5e-4$, decaying by a factor of $0.5$ every $10$ epochs. the value of nearest neighbor $k$ is 3.
Our model is trained end-to-end, so fine-tuning during the testing phase is not necessary.In the testing phase, we randomly constructed $10000$ episodes from the test datasets to calculate the final classification results. 
Additionally, we adopt the top-1 mean accuracy criterion and report $95\%$ confidence interval to ensure the reliability of the experimental results.

\begin{table}[]
\caption{
5-way 1-shot and 5-shot classification accuracies on CUB-200 dataset using Conv-4 and ResNet-12 backbones with $25\%$ confidence intervals. All the results of comparative methods are from the exiting literature ('-' not reported).}
\label{tab2}
\resizebox{\linewidth}{!}{
\begin{tabular}{lcccc}
\toprule
\multicolumn{1}{c}{\multirow{2}{*}{\textbf{Method}}} & \multicolumn{2}{c}{\textbf{Conv-4}}    & \multicolumn{2}{c}{\textbf{ResNet-12}}       \\ \cmidrule(r){2-3} \cmidrule(r){4-5}
\multicolumn{1}{c}{} & 1-shot & 5-shot & 1-shot & 5-shot \\ \hline
ProtoNet\cite{snell2017prototypical}  & 63.73  & 81.50  & 66.09  & 82.50  \\
DN4\cite{li2019distribution}   & 73.42  & 90.38  & -   & -   \\
DSN\cite{simon2020adaptive}  & 66.01  & 85.41  & 80.80  & 91.19  \\
CTX\cite{doersch2020crosstransformers}   & 69.64  & 87.31  & 78.47  & 90.90  \\
DeepEMD\cite{zhang2020deepemd}  & -  & -  & 77.14  & 88.98  \\
FRN\cite{wertheimer2021few}   & 73.48  & 88.43  & 83.16  & 92.59  \\
DMN4\cite{liu2022dmn4}  & 78.36  & 92.16  & -  & - \\ \hline
\textbf{TALDS-Net(ours)}  & \multicolumn{1}{l}{\textbf{81.73}} & \multicolumn{1}{l}{\textbf{93.24}} & \multicolumn{1}{l}{\textbf{88.32}} & \multicolumn{1}{l}{\textbf{94.90}} \\ 
\bottomrule
\end{tabular}
}
\end{table}
\subsection{Comparisons with State-of-the-art Methods}
\textbf{Results on miniImageNet dataset.}
As shown in Table \ref{tab1}, our method outperforms the SOTA results in both 5-way 1-shot and 5-shot tasks. 
When using Conv-4 as the backbone, compared to DN4 and DMN4, our method shows significant improvements. In the 5-way 1-shot task, it achieves a gain of $5.54\%$ and $1.01\%$, respectively. 
In the 5-way 5-shot task, it achieves a gain of $3.61\%$ and $0.41\%$, respectively. Compared to the state-of-the-art (SOTA), our method improves by $0.9\%$ in 5-way 1-shot task and $0.41\%$ in 5-way 5-shot task. 
Similarly, when using ResNet-12 as the backbone, compared to DN4 and DMN4, our method achieves improvements of $2.54\%$ and $1.31\%$ in the 5-way 1-shot task, and $3.21\%$ and $0.79\%$ in the 5-way 5-shot task, respectively. 
Compared to SOTA, our method improves by $1.31\%$ and $0.79\%$ in the two settings, respectively.

\textbf{Results on tieredImageNet dataset.}
As shown in Table \ref{tab1}, our method outperforms the SOTA methods when using Conv-4 as the backbone. Compared to DN4 and DMN4, our approach achieves improvements of $4.65\%$ and $0.55\%$ in the 5-way 1-shot task, and $3.61\%$ and $0.41\%$ in the 5-way 5-shot setting, respectively. When using ResNet-12 as the backbone, our method surpasses DN4 by $1.74\%$ in 5-way 1-shot task and $2.71\%$ in 5-shot task. Compared to DMN4, it achieves a $0.4\%$ improvement in the 5-way 5-shot task but ranks second to the current SOTA method.

\textbf{Results on fine-grained CUB-200 dataset.}
As shown in Table \ref{tab2}, our method indicates the effectiveness on fine-grained datasets. When using Conv-4 as the backbone, our method outperforms the current best methods by $3.37\%$ and $1.08\%$ in the 5-way 1-shot and 5-shot tasks, respectively. When using ResNet-12 as the backbone, our method surpasses the current best methods by $5.16\%$ and $2.31\%$ in the same settings.

\begin{table}
\caption{Ablation study on miniImageNet and CUB-200 datasets for the TALDS-Net.}
\label{tab3}
\resizebox{\linewidth}{!}{
\small
\begin{tabular}{ccccccc}
\toprule
\multicolumn{1}{c}{\multirow{2}{*}{$\mathcal{F}_\Gamma$}} & \multicolumn{1}{c}{\multirow{2}{*}{$\mathcal{F}_\Psi$}} & \multicolumn{2}{c}{miniImageNet} & \multicolumn{2}{c}{CUB-200} \\ \cmidrule(l){3-4} \cmidrule(l){5-6}
& \multicolumn{1}{c}{} & 1-shot & 5-shot & 1-shot & 5-shot \\ 
\midrule
\ding{55} & \ding{55} & 51.20 & 71.12 & 73.42 & 89.20 \\
\ding{55} & \checkmark & 54.31 & 73.71 & 79.01 & 93.36 \\
\checkmark & \ding{55} & 53.48 & 72.65 & 75.93 & 89.57 \\
\checkmark & \checkmark & \textbf{56.78} & \textbf{74.63} & \textbf{81.32} &\textbf{93.24} \\ 
\bottomrule
\end{tabular}
}
\end{table}

\subsection{Ablation Study}
\textbf{Analysis on modules of the proposed method.} In this section, we primarily discuss the necessity of adaptively selecting query and support descriptors. First, to validate the effectiveness of adaptively selecting support descriptors and query descriptors over using all support descriptors, we conducted ablation experiments using Conv-4 as the backbone on the miniImageNet dataset, and the results are shown in Table \ref{tab3}. We removed $\mathcal{F}_{\Gamma}$ and $\mathcal{F}_{\Psi}$ from the network to ensure that each part is independent. It can be seen that the module $\mathcal{F}_{\Gamma}$ for adaptive query descriptor selection has a significant impact on performance improvement. It improved by $3.11\%$ and $2.59\%$ on miniImageNet and by $5.59\%$ and $4.16\%$ on CUB-200, respectively. The module $\mathcal{F}_{\Psi}$ for adaptive support descriptor selection also improved performance with an increase of $2.28\%$, $1.53\%$, $2.51\%$, and $0.37\%$, respectively. And even when we use only $\mathcal{F}_{\Psi}$, our method outperforms DN4 and DMN4. 
So, $\mathcal{F}_{\Gamma}$ and $\mathcal{F}_{\Psi}$ are effective for the overall method.


\section{CONCLUSIONS}
\label{sec:conclusions}

In this paper, we propose a novel Task-Aware Adaptive local descriptors selection Network (TALDS-Net) for Few-shot Image Classification, aiming to learn discriminative local representations in support and query images. Our two task-aware adaptive modules efficiently select discriminative local descriptors for specific tasks. We hope that our method can provide meaningful insights for other researchers in terms of deep local descriptors.

\textbf{Acknowledgement}: This work is supported by National Key R\&D Program of China (2018YVFA0701700, 2018YFA0701701), NSFC (62176172, 61672364).

\vfill\pagebreak

\bibliographystyle{IEEEbib}
\clearpage
\bibliography{refs}

\end{document}